\pdfoutput=1
\documentclass[11pt]{article}

\usepackage[final]{emnlp2021}
\usepackage{times}
\usepackage{latexsym}
\usepackage{multirow}
\usepackage{booktabs}
\usepackage{graphicx}
\usepackage{amssymb}
\usepackage{makecell}
\usepackage{amsmath, bm}
\usepackage{longtable}
\usepackage{subfigure}
\usepackage{caption}
\usepackage{enumerate}
\usepackage{enumitem}
\usepackage[T1]{fontenc}
\usepackage[utf8]{inputenc}
\usepackage{microtype}

\title{Fine-grained Entity Typing via Label Reasoning}

\author{
  Qing Liu${}^{1,3}$,
  Hongyu Lin${}^{1}$\thanks{~ Corresponding authors.},
  Xinyan Xiao${}^{4}$,
  Xianpei Han${}^{1,2}$\footnotemark[1],
  Le Sun${}^{1,2}$,
  Hua Wu${}^{4}$
  \\
  ${}^{1}$Chinese Information Processing Laboratory ~
  ${}^{2}$State Key Laboratory of Computer Science \\
  Institute of Software, Chinese Academy of Sciences, Beijing, China\\
  ${}^{3}$University of Chinese Academy of Sciences, Beijing, China \\
  ${}^{4}$Baidu Inc., Beijing, China \\
  {\tt \{liuqing2020,hongyu,xianpei,sunle\}@iscas.ac.cn} \\
 {\tt \{xiaoxinyan,wu\_hua\}@baidu.com}
}

\begin{document}
\maketitle
\begin{abstract}
Conventional entity typing approaches are based on independent classification paradigms, which make them difficult to recognize inter-dependent, long-tailed and fine-grained entity types. In this paper, we argue that the implicitly entailed extrinsic and intrinsic dependencies between labels can provide critical knowledge to tackle the above challenges. To this end, we propose \emph{Label Reasoning Network(LRN)}, which sequentially reasons fine-grained entity labels by discovering and exploiting label dependencies knowledge entailed in the data. Specifically, LRN utilizes an auto-regressive network to conduct deductive reasoning and a bipartite attribute graph to conduct inductive reasoning between labels, which can effectively model, learn and reason complex label dependencies in a sequence-to-set, end-to-end manner. Experiments show that LRN achieves the state-of-the-art performance on standard ultra fine-grained entity typing benchmarks, and can also resolve the long tail label problem effectively.

\end{abstract}
\section{Introduction}
Fine-grained entity typing (FET) aims to classify entity mentions to a fine-grained semantic label set, e.g., classify ``\textit{FBI agents}" in ``\textit{They were arrested by \textbf{FBI agents}.}" as \{\textit{organization, administration, force, agent, police}\}. By providing fine-grained semantic labels, FET is critical for entity recognition~\citep{DBLP:conf/acl/LinLHS19_icip1, DBLP:conf/emnlp/LinLHSDJ19_icip2, DBLP:conf/emnlp/LinLTHSWY20_icip3, DBLP:conf/aaai/ZhangLH0LWY21_icip4, zhang-etal-2021-de} and can benefit many NLP tasks, such as relation extraction~\cite{DBLP:conf/eacl/SchutzeYA17_RE_downstream1, DBLP:conf/acl/ZhangHLJSL19_RE_downstream2}, entity linking~\cite{DBLP:conf/aaai/OnoeD20_EL_downstream4} and question answering~\cite{DBLP:conf/emnlp/YavuzGSSY16_QA_downstream3}.
\begin{figure}[htb!]
\setlength{\belowcaptionskip}{-0.5cm}
\centering
	\subfigure[Extrinsic dependency]{
	\resizebox{0.22\textwidth}{!}{
			\includegraphics{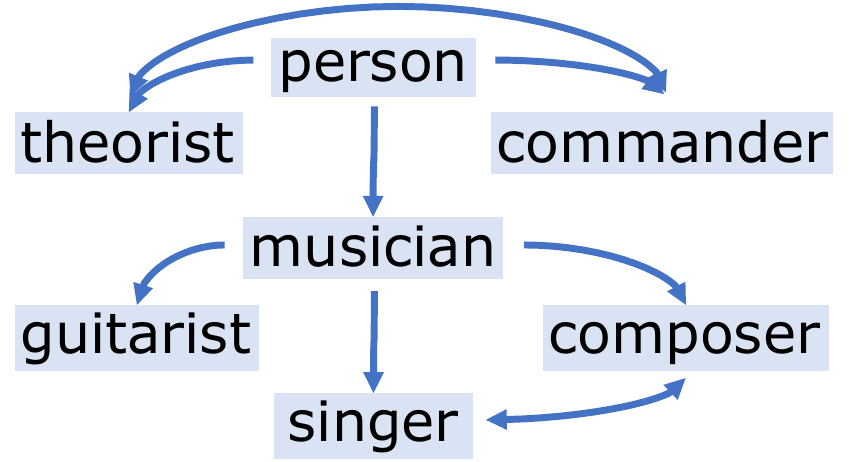}
		\label{Fig.introduction(a)}
	    }}
	\subfigure[Intrinsic dependency]{
	\resizebox{0.22\textwidth}{!}{
   	 	\includegraphics{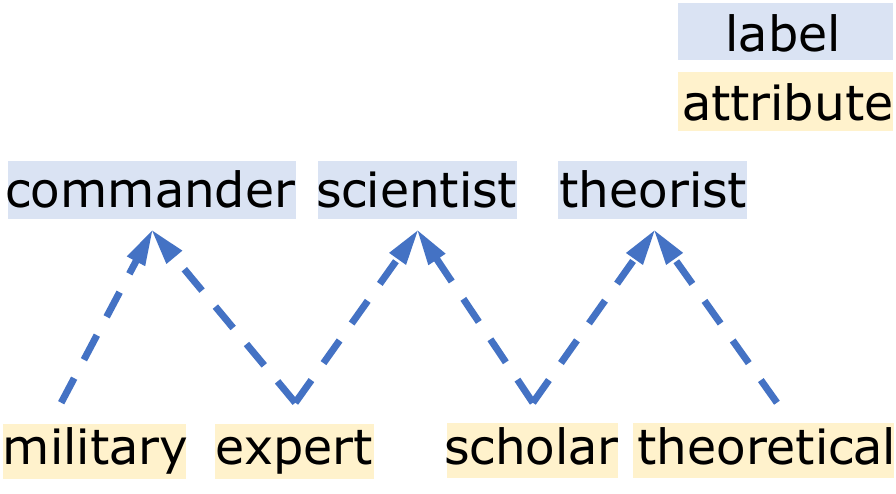}
	    \label{Fig.introduction(b)}
	    }}
	\subfigure[Label reasoning process]{
	 \resizebox{0.48\textwidth}{!}{
   	 	\includegraphics{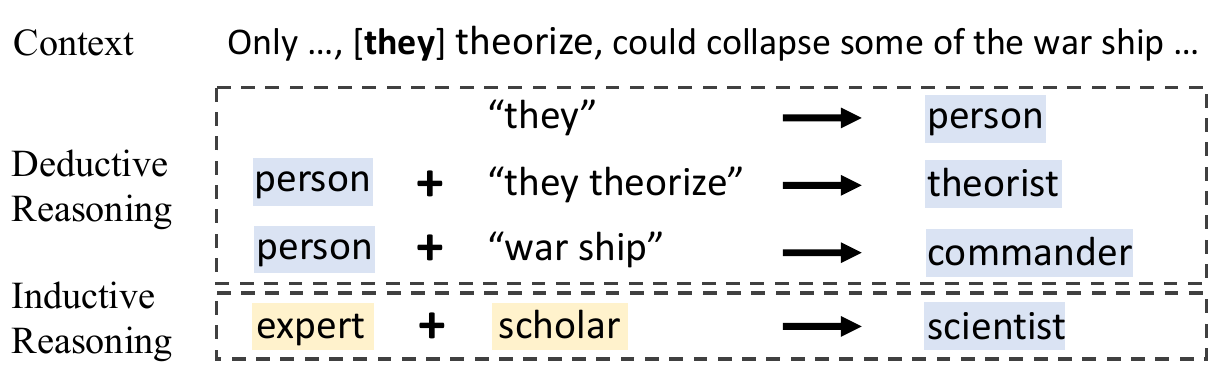}
	    \label{Fig.introduction(c)}
	    }}
	\caption{Examples of deductive reasoning based on the extrinsic dependency and inductive reasoning based on the intrinsic dependency, where the labels \textit{person}, \textit{theorist} and \textit{commander} are deducted respectively and the label \textit{scientist} is inducted from the attributes \{\texttt{expert}, \texttt{scholar}\}.}
	\label{Fig.introduction}
\end{figure}

The fundamental challenge of FET comes from its large-scale and fine-grained entity label set, which leads to significant difference between FET and conventional entity typing. First, due to the massive label set, it is impossible to independently recognize each entity label without considering their dependencies. For this, existing approaches use the predefined label hierarchies~\citep{ren2016afet_hier1, DBLP:conf/eacl/InuiRSS17_shimaoka_hier7, DBLP:conf/eacl/AbhishekAA17_hier5, DBLP:conf/eacl/SchutzeWK17_hier8, XuandBarbosa2018neural_hier2, DBLP:conf/ijcai/WuZMGH19_hier3, DBLP:conf/acl/ChenCD20_onto_hier4, ren2020fine_inference_hier6} or label co-occurrence statistics from training data~\citep{DBLP:conf/acl/RabinovichK17_core1, xiong2019imposing_core2, linandJi2019attentive_core3} as external constraints. Unfortunately, these label structures or statistics are difficult to obtain when transferring to new scenarios. Second, because of the fine-grained and large-scale label set, many long tail labels are only provided with several or even no training instances. For example, in Ultra-Fine dataset~\citep{choi2018ultra_data1}, $>$80\% of entity labels are with $<$5 instances, and more seriously 25\% of labels never appear in the training data. However, training data can provide very limited direct information for these labels, and therefore previous methods commonly fail to recognize these long-tailed labels.

Fortunately, the implicitly entailed label dependencies in the data provide critical knowledge to tackle the above challenges. Specifically, the dependencies between labels exist extrinsically or intrinsically. On the one hand, the extrinsic dependencies reflect the \emph{direct} connections between labels, which partially appear in the form of label hierarchy and co-occurrence. For example, in Figure~\ref{Fig.introduction(a)} the labels \textit{person}, \textit{musician}, \textit{composer} are with extrinsic dependencies because they form a three-level taxonomy. Furthermore, \textit{singer} and  \textit{composer} are also with extrinsic dependency because they often co-occur with each other. On the other hand, the intrinsic dependencies entail the \emph{indirect} connections between labels through their underlying attributes. For the example in Figure~\ref{Fig.introduction(b)}, label \textit{theorist} and \textit{scientist} share the same underlying attribute of \texttt{scholar}. Such intrinsic dependencies provide an effective way to tackle the long tail labels, because many long tail labels are actually composed by non-long tail attributes which can be summarized  from non-long tail labels.

To this end, this paper proposes \textit{Label Reasoning Network (LRN)}, which uniformly models, learns and reasons both extrinsic and intrinsic label dependencies without given any predefined label structures. Specifically, LRN utilizes an auto-regressive network to conduct deductive reasoning and a bipartite attribute graph to conduct inductive reasoning between labels. Both of these two kinds of mechanisms are jointly applied to sequentially generate fine-grained labels in an end-to-end, sequence-to-set manner. Figure~\ref{Fig.introduction(c)} shows several examples. To capture extrinsic dependencies, LRN introduces deductive reasoning (i.e., draw a conclusion based on premises) between labels, and formulates it using an auto-regressive network to predict labels based on both the context and previous labels. For example, given previously-generated label \textit{person} of the mention \textit{they}, as well as  the context \textit{they theorize}, LRN will deduce its new label \textit{theorist} based on the extrinsic dependency between \textit{person} and \textit{theorist} derived from data. For intrinsic dependencies, LRN introduces inductive reasoning (i.e., gather generalized information to a conclusion), and utilizes a bipartite attribute graph to reason labels based on current activated attributes of previous labels. For example, if the attributes \{\texttt{expert}, \texttt{scholar}\} have been activated, LRN will induce a new label \textit{scientist} based on the attribute-label relations. Consequently, by decomposing labels into attributes and associating long tail labels with frequent labels, LRN can also effectively resolve the long tail label problem by leveraging their non-long tail attributes. Through jointly leveraging the extrinsic and intrinsic dependencies via deductive and inductive reasoning, LRN can effectively handle the massive label set of FET.

Generally, our main contributions are:
\begin{itemize}[leftmargin=0.6cm,topsep=0.1cm]
\setlength{\itemsep}{0cm}
\setlength{\parskip}{0.1cm}
\item We propose \textit{Label Reasoning Network}, which uniformly models, automatically learns and effectively reasons the complex dependencies between labels in an end-to-end manner.

\item To capture extrinsic dependencies, LRN utilizes deductive reasoning to sequentially reason labels via an auto-regressive network. In this way, extrinsic dependencies are discovered and exploited without predefined label structures.

\item To capture intrinsic dependencies, LRN utilizes inductive reasoning to reason labels via a bipartite attribute graph. By decomposing labels into attributes and associating long-tailed labels with frequent attributes, LRN can effectively reason long-tailed and even zero-shot labels.
\end{itemize}

We conduct experiments on standard Ultra-Fine~\citep{choi2018ultra_data1} and OntoNotes~\citep{ontonotes_data2} dataset. Experiments show that our method achieves new state-of-the-art performance: a 13\% overall F1 improvement and a 44\% F1 improvement in the ultra-fine granularity.\footnote{Our source codes are openly available at \href{https://github.com/loriqing/Label-Reasoning-Network}{https://github.com/loriqing/Label-Reasoning-Network}}

\begin{figure*}[tbh!]
\setlength{\belowcaptionskip}{-0cm}
\centering
\includegraphics[width=0.88\textwidth]{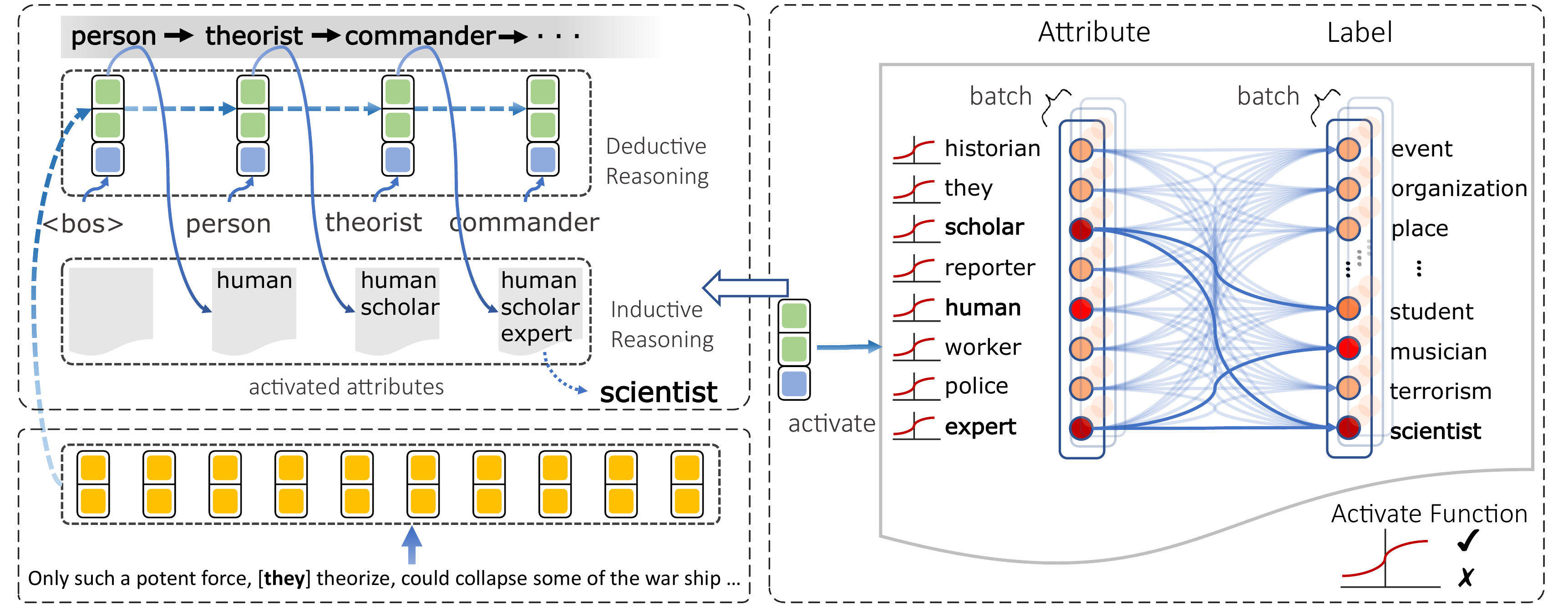}
\caption{Overview of the process for LRN which contains an encoder, a deductive reasoning-based decoder and an inductive reasoning-based decoder. The figure shows: at step 1, the label \textit{person} is predicted by deductive reasoning, and the attribute \texttt{human} is activated; at step 3, the label \textit{scientist} is generated by inductive reasoning.}
\label{Fig.framework}
\end{figure*}

\section{Related Work}
One main challenge for FET is how to exploit complex label dependencies in the large-scale label set. Previous studies typically use predefined label hierarchy and co-occurrence structures estimated from data to enhance the models. To this end, \citet{ren2016afet_hier1, XuandBarbosa2018neural_hier2, DBLP:conf/ijcai/WuZMGH19_hier3, DBLP:conf/acl/ChenCD20_onto_hier4} design new loss function to exploit label hierarchies. \citet{DBLP:conf/eacl/AbhishekAA17_hier5} enhance the label representation by sharing parameters. \citet{DBLP:conf/eacl/InuiRSS17_shimaoka_hier7, McCallumVVMR18_space1, DBLP:conf/emnlp/Lopez020_hyper} embed labels into a high-dimension or a new space. And the studies exploit co-occurrence structures including limiting the label range during label set prediction~\citep{DBLP:conf/acl/RabinovichK17_core1}, enriching the label representation by introducing associated labels~\citep{xiong2019imposing_core2}, or requiring latent label representation to reconstruct the co-occurrence structure~\citep{linandJi2019attentive_core3}. However, these methods require predefined label structures or statistics from training data, and therefore is difficult to be extended to new entity types or domains.

The ultra fine-grained label set also leads to data bottleneck and the long tail problem. In recent years, some previous approaches try to tackle this problem by introducing zero/few-shot learning methods~\citep{ma2016label_few1, HuangMPJ16_open1, ZhouKTR18_open2, YuanD18_few2, obeidat2019description_few3, zhang2020mzet_few4, DBLP:conf/www/RenLZ20_zero}, or using data augmentation with denosing strategies~\citep{DBLP:conf/kdd/RenHQVJH16_denoise3, DBLP:conf/naacl/OnoeD19_elmo, DBLP:conf/ijcai/ZhangLXZXHW20_denoise1, DBLP:conf/aaai/AliSLW20_denoise2} or utilizing external knowledge~\citep{DBLP:conf/emnlp/CorroAGW15_KB, DBLP:conf/emnlp/DaiDLS19_el} to introduce more external knowledge.

In this paper, we propose Label Reasoning Network, which is significantly different from previous methods because 1) by introducing deductive reasoning, LRN can capture extrinsic dependencies between labels in an end-to-end manner without predefined structures; 2) by introducing inductive reasoning, LRN can leverage intrinsic dependencies to predict long tail labels; 3) Through the sequence-to-set framework, LRN can consider two kinds of label dependencies simultaneously to jointly reason frequent and long tail labels.

\section{Label Reasoning Network for FET}
Figure~\ref{Fig.framework} illustrates the framework of \textit{Label Reasoning Network}. First, we encode entity mentions through a context-sensitive encoder, then sequentially generate entity labels via two label reasoning mechanisms: deductive reasoning for exploiting extrinsic dependencies and inductive reasoning for exploiting intrinsic dependencies. In our Seq2Set framework, the label dependency knowledge can be effectively modeled in the parameters of LRN, automatically learned from training data, and naturally exploited during the sequential label decoding process. In the following we describe these components in detail. 

\subsection{Encoding}
For encoding, we form the input instance $\mathcal{X}$ as ``[CLS], $x_1$, ..., [$E_1$], $m_1$, ..., $m_k$, [$E_2$], ..., $x_n$" where [$E_1$], [$E_2$] are entity markers, $m$ is mention word and $x$ is context word. We then feed $\mathcal{X}$ to BERT and obtain the source hidden state $\mathcal{H}=\{h_1,...,h_n\}$. Finally, the hidden vector of [CLS] token is used as sentence embedding $\bm{g}$.

\subsection{Deductive Reasoning for Extrinsic Dependencies}
This section describes how to capture extrinsic dependencies for label prediction via a deductive reasoning mechanism. To this end, the deductive reasoning-based decoder sequentially generates labels based on both context and previous labels, e.g., ``\textit{for his books}" + \textit{person} $\rightarrow$ \textit{writer} and ``\textit{record an album}" + \textit{person} $\rightarrow$ \textit{musician}. In this way, a label is decoded by considering both context-based prediction and previous labels-based prediction.

Concretely, we utilize a LSTM-based auto-regressive network as decoder and obtain the hidden state of decoder $\mathcal{S}=\{s_0,...,s_{k}\}$, where $k$ is the number of predicted labels. We first initialize $s_0$ using sentence embedding $\bm{g}$, then at each time step, two attention mechanisms -- contextual attention and premise attention, are designed to capture context and label information for next prediction.

\paragraph{Contextual Attention} is used to capture the context evidence for label prediction. For example, the context ``\textit{they theorize}" provides rich information for \textit{theorist} label. Specifically, at each time step $t$, contextual attention identifies relevant context by assigning a weight $\alpha_{ti}$ to each $\bm{h}_i$ in the source hidden state $\mathcal{H}$: 
\begin{align}
e_{ti} &= \bm{v}_c^{T}tanh(\bm{W}_c \bm{s}_t + \bm{U}_c \bm{h}_i)  \label{att1_e}
\\
\alpha_{ti} &= \frac{exp(e_{ti})}{\sum_{i=1}^{n}exp(e_{ti})}  \label{att1_a}
\end{align}
where $\bm{W}_c$, $\bm{U}_c$, $\bm{v}_c$ are weight parameters and $\bm{s}_t$ is the hidden state of decoder at time step $t$. Then the context representation $\bm{c}_t$ is obtained by:
\begin{align}
\bm{c}_t &= \sum_{i=1}^{n}\alpha_{ti}\bm{h}_i  \label{att1_c}
\end{align}
\paragraph{Premise Attention} exploits the dependencies between labels for next label prediction. For example, if \textit{person} has been generated, its hyponym label \textit{theorist} will be highly likely to be generated in context ``\textit{they theorize}". Concretely, at each time step $t$, premise attention captures the dependencies to previous labels by assigning a weight $\alpha_{tj}$ to each $\bm{s}_j$ of previous hidden states of decoder $\mathcal{S}_{<t}$:
\begin{align}
e_{tj} &= \bm{v}_p^{T}tanh(\bm{W}_p \bm{s}_t + \bm{U}_p\bm{s}_j) \label{att2_e} \\  
\alpha_{tj} &= \frac{exp(e_{tj})}{\sum_{j=0}^{t-1}exp(e_{tj})}  \label{att2_a}
\end{align}
where $\bm{W}_p$, $\bm{U}_p$, $\bm{v}_p$ are weight parameters. Then the previous label information $\bm{u}_t$ is obtained by:
\begin{align}
\bm{u}_t &= \sum_{j=0}^{t-1}\alpha_{tj}\bm{s}_j  \label{att1_u}
\end{align}
\paragraph{Label Prediction. }Given the context representation $\bm{c}_t$ and the previous label information $\bm{u}_t$, we use $\bm{m}_t=[\bm{c}_t+\bm{g}; \bm{u}_t+\bm{s}_t]$ as input, and calculate the probability distribution over label set $L$:
\begin{align}
\bm{s}_t &= {\rm{LSTM}}(\bm{s}_{t-1}, \bm{W}_b\bm{y}_{t-1}) \\
\bm{o}_t &= \bm{W}_o\bm{m}_t \\
\bm{y}_t &= softmax(\bm{o}_t+\bm{I}_t)
\end{align}
where $\bm{W}_o$ and $\bm{W}_b$ are weight parameters and we use the mask vector $\bm{I}_t \in \mathbb{R}^{L+1}$ \cite{YangSLMWW18_sgm} to prevent duplicate predictions.
\begin{align}
(\bm{I}_t)_i = \begin{cases}
-\inf &, l_i \in \mathcal{Y}^*_{t-1} \\
1 &, {\rm otherwise} \\
\end{cases}
\end{align}
where $\mathcal{Y}^*_{t-1}$ is the predicted labels before step $t$ and $l_i$ is the $i^{th}$ label in label set $L$. The label with maximum value in $\bm{y}_t$ is generated and used as the input for the next time step until $[EOS]$ is generated.
\subsection{Inductive Reasoning for Intrinsic Dependencies}
Deductive reasoning can effectively capture extrinsic dependencies. However, labels can also have intrinsic dependencies if they share attributes, e.g., \textit{theorist} and \textit{scientist} shares \texttt{scholar} attribute. To leverage intrinsic dependencies, LRN conducts inductive reasoning by associating labels to attributes via a bipartite attribute graph. A label will be generated if most of its attributes are activated. Instead of heuristically setting the number of attributes to be activated, we select labels based on their overall activation score from all attributes. By capturing such label-attribute relations, many long tail labels can be effectively predicted because they are usually related to non-long tail attributes. 

To this end, we first design a bipartite attribute graph to represent attribute-label relations. Based on the bipartite attribute graph, at each time step, attributes will be activated based on the hidden state of decoder, and new labels will be inducted by reasoning over the activated attributes. For example, in Figure~\ref{Fig.framework} the predicted labels \textit{person}, \textit{theorist} and \textit{commander} will correspondingly activate the attributes \texttt{human}, \texttt{scholar} and \texttt{expert}, and then the \textit{scientist} label will be activated via inductive reasoning based on these attributes.

\paragraph{Bipartite Attribute Graph (BAG).}\label{AG method} BAG $\mathcal{G}=\{V,E\}$ is designed to capture the relations between attributes and labels. Specifically, nodes $V$ contain attribute nodes $V_a$ and label nodes $V_l$, and edges $E$ only exist between attributes nodes and labels nodes, with the edge weight indicating the attribute-label relatedness. Attributes are represented using natural language words in BAG. Figure~\ref{Fig.framework} shows a BAG where $V_a$ contains words \{\texttt{scholar}, \texttt{expert}, \texttt{historian}, ...\}, $V_l$ are all entity labels in label set $\textit{L}$, containing \{\textit{student}, \textit{musician}, \textit{scientist}, ...\}

\begin{figure}[!t]
\vspace{-0cm}
\setlength{\belowcaptionskip}{-0.3cm}
\centering
\includegraphics[width=0.48\textwidth]{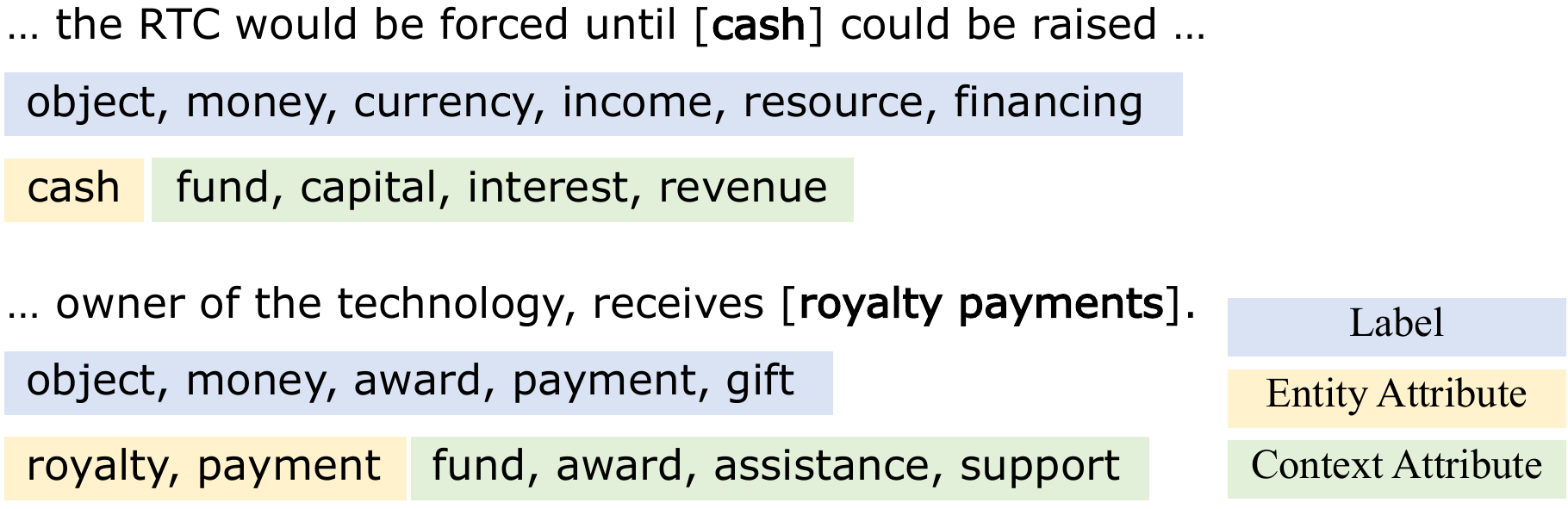} 
\caption{Examples of attributes.}
\label{Fig.attributes} 
\end{figure}

\paragraph{BAG Construction.}\label{collection} Because there are many labels and many attributes, we dynamically build a local BAG during the decoding for each instance. In this way the BAG is very compact and the computation is very efficient~\citep{DBLP:journals/ai/ZupanBDB99_attribute_reduce}. In local BAG, we collect attributes in two ways: (1) We mask the entity mention in the sentence, and predict the [MASK] token using masked language model (this paper uses BERT-base-uncased), and the non-stop words whose prediction scores greater than a confidence threshold ${\theta}_c$ will be used as attributes --- we denote them as context attributes; Since PLM usually predicts high-frequency words, the attributes are usually not long-tailed, which facilitates modeling dependencies between head and tail labels. This mask-prediction strategy is also used in~\citet{DBLP:conf/emnlp/XinZH0S18_put_back}, for collecting additional semantic evidence of entity labels. (2) We directly segment the entity mention into words using Stanza\footnote{https://pypi.org/project/stanza/}, and all non-stop words are used as attributes --- we denote them as entity attributes. Figure~\ref{Fig.attributes} shows several attribute examples. Given attributes, we compute the attribute-label relatedness (i.e. $E$ in $\mathcal{G}$) using the cosine similarity between their GloVe embeddings~\citep{DBLP:conf/emnlp/PenningtonSM14_glove}. 

\paragraph{Reasoning over BAG.} At each time step, we activate attributes in BAG by calculating their similarities to the current hidden state of decoder $\bm{s}_t$. For the $i^{th}$ attribute node ${V_a}^{(i)}$, its activation score is:
\begin{align}
{score}_{V_a}^{(i)} &= ReLU(sim(\bm{W}_s \bm{s}_t, \bm{W}_a {V_a}^{(i)})
\end{align}
where $\bm{W}_s$ is the weight parameter, $\bm{W}_a$ is the attribute embedding (i.e., word embedding of attribute words). We use cosine distance to measure similarity and employ ReLU to activate attributes. Then we induce new labels by reasoning over the activated attributes as:
\begin{align}
{score}_{V_l}^{(j)} &= \sum_{i=1}^{n_a} {score}_{V_a}^{(i)} E_{ij}
\end{align}
where $n_a$ is the number of attributes, ${V_l}^{(j)}$ is the $j_{th}$ label nodes and $E_{ij}$ is the weight between them. Finally a label will be generated if its activation score is greater than a similarity threshold ${\theta}_s$.

Note that our inductive reasoning and deductive reasoning are jointly modeled in the same decoder, i.e., they share the same decoder hidden state  but with different label prediction process. Once deductive reasoning-based decoder generates \textit{[EOS]}, the label prediction stops. Finally, we combine the predicted labels of both deductive reasoning and inductive reasoning as the final FET results.

\section{Learning}
In FET, each instance is represented as \{$\mathcal{X}$, $\mathcal{Y}$\} where $\mathcal{X}$ is ``[CLS], $x_1$, ..., [$E_1$], $m_1$, ..., $m_k$, [$E_2$],..., $x_n$" and $\mathcal{Y}=\{y_1,...,y_m\}$ is the golden labels. To learn our model, we design two losses: set prediction loss for deductive reasoning-based decoding and BAG loss for inductive reasoning-based decoding.

\paragraph{Set Prediction Loss.}
In FET, cross entropy loss is not appropriate because the prediction results is a label set, i.e., \{$y^{*}_1$, $y^{*}_2$, $y^{*}_3$\} and \{$y^{*}_3$, $y^{*}_2$, $y^{*}_1$\} should have the same loss. Therefore we measure the similarity of two label set using the bipartite matching loss~\citep{DBLP:journals/corr/abs-2011-01675_matching_loss}. Given the golden label set $\mathcal{Y}=\{y_1,...,y_m\}$ and generated label set ${\mathcal{Y}}^{*}=\{{y^{*}_1},...,{y^{*}_m}\}$, the matching loss $\mathcal{L}(ij)_{S}$ of $y_i$ and ${y^{*}_j}$ is calculated by \ref{single_loss}, then we use the Hungarian Algorithm~\citep{kuhn1955hungarian_loss} to get the specific order of golden label set as $\widetilde{\mathcal{Y}}=\{\widetilde{y}_1,...,\widetilde{y}_m\}$ to obtain minimum matching loss $\mathcal{L}_{S}$:
\begin{align}
\mathcal{L}(ij)_{S} &= \textup{CE}(y_i, y^{*}_j) \label{single_loss}  \\
\mathcal{L}_{S} &= \textup{CE}(\widetilde{\mathcal{Y}}, \mathcal{Y}^{*}) \label{total_loss} 
\end{align}
where \textup{CE} is cross-entropy. 

\paragraph{BAG Loss.}
To make the model activate labels correctly, we add a supervisory loss to the bipartite attribute graph to active correct labels:
\begin{align}
\mathcal{L}_{A} &= -\sum_{j=1}^{|\textit{L}|}{score}_{V_l}^{(j)} * y_j \\
y_j &= \begin{cases}
1 &,  v_j \in \mathcal{Y} \\
-1 &, v_j \notin \mathcal{Y} 
\end{cases} 
\end{align}\normalsize
\paragraph{Final Loss.} The final loss is a combination of set loss and BAG loss:
\begin{align}
    \mathcal{L} = \mathcal{L}_{S} + \lambda \mathcal{L}_{A}
\end{align}
where $\lambda$ is the relative weight of these two losses\footnote{ In our auxiliary experiments, we find that its impact is minor, so this paper empirically sets it to 1.}.
\section{Experiments}
\subsection{Settings}
\paragraph{Datasets} We conduct experiments on two standard fine-grained entity typing datasets\footnote{Released in https://github.com/uwnlp/open\_type}: Ultra-Fine as primary dataset and OntoNotes as complementary dataset. Ultra-Fine contains 6K manually-annotated examples, 2519 categories, and 5.4 labels per sample on average. Followed~\citet{choi2018ultra_data1} we use the same 2K/2K/2K train/dev/test splits and evaluate using macro precision, recall and F-score. Original OntoNotes dataset~\citep{ontonotes_data2} contains 25K/2K/9K train/dev/test data, 89 categories and 2.7 labels per sample on average. And \citet{choi2018ultra_data1} offers an augmentation training data with 2.3 labels per sample on average. We evaluate on both versions using the standard metrics: accuracy, macro F-score and micro F-score.
\paragraph{Baselines} For Ultra-Fine dataset, we compare with following baselines: \citet{DBLP:conf/naacl/OnoeD19_elmo} which offers two multi-classifiers using BERT and ELMo as encoder respectively, \citet{choi2018ultra_data1} which is a multi-classifier using GloVe+LSTM as encoder, \citet{xiong2019imposing_core2} which is a multi-classifier using GloVe+LSTM as encoder and exploits label co-occurrence via introducing associated labels to enrich the label representation, \citet{DBLP:conf/emnlp/Lopez020_hyper} which is a hyperbolic multi-classifier using GloVe. For OntoNotes dataset, in addition to the baselines for Ultra-Fine, we also compare with \citet{wang2020empirical} which offers a multi-classifier using BERT as encoder, \citet{linandJi2019attentive_core3} which offers a multi-classifier using ELMo as encoder and exploits label co-occurrence via requiring the latent representation to reconstruct the co-occurrence association and \citet{DBLP:conf/acl/ChenCD20_onto_hier4} which offers a multi-classifier using ELMo as encoder and exploits label hierarchy via designing a hierarchy-aware loss function.
\paragraph{Implementation} We use BERT-Base(uncased) \citep{DBLP:conf/naacl/DevlinCLT19_bert} as encoder, Adam optimizer \citep{DBLP:journals/corr/KingmaB14_adam} with learning rate of BERT as 5e-5 and of other parameters as 1e-3. The batch size is 32, encoder hidden size is 768, the decoder hidden size is 868 and label embedding size is 100, the dropout rate of decoder is 0.6. The confidence threshold ${\theta}_c$ and the similarity threshold ${\theta}_s$ both are optimized on dev set and set as 0.1 and 0.2 respectively.
 We use the GloVe embedding \citep{DBLP:conf/emnlp/PenningtonSM14_glove} to represent the nodes of BAG and fix it while training.
\subsection{Overall Results}
\begin{table}[!t]
\setlength{\belowcaptionskip}{-0.4cm}
\centering
\resizebox{.45\textwidth}{!}{
\begin{tabular}{lccc}
\Xhline{1.2pt}
\multicolumn{1}{l|}{\textbf{Model}} & \textbf{P} & \textbf{R} & \textbf{F1} \\ \hline
\multicolumn{4}{c}{without label dependency} \\ \hline
\multicolumn{1}{l|}{*\small\citet{choi2018ultra_data1}\normalsize} & 47.1 & 24.2 & 32.0 \\
\multicolumn{1}{l|}{*ELMo\small\citep{DBLP:conf/naacl/OnoeD19_elmo}\normalsize} & 51.5 & 33.0 & 40.2 \\
\multicolumn{1}{l|}{BERT\small\citep{DBLP:conf/naacl/OnoeD19_elmo}\normalsize} & 51.6 & 33.0 & 40.2 \\
\multicolumn{1}{l|}{BERT{[}in-house{]}} & 55.9 & 33.0 & 41.5 \\ \hline
\multicolumn{4}{c}{with label dependency} \\ \hline
\multicolumn{1}{l|}{*L\small{ABEL}\normalsize{GCN} \small\citep{xiong2019imposing_core2}\normalsize} & 50.3 & 29.2 & 36.9 \\
\multicolumn{1}{l|}{LRN \small w/o IR \normalsize} & \textbf{61.2} & 33.5 & 43.3 \\
\multicolumn{1}{l|}{LRN} & 54.5 & \textbf{38.9} & \textbf{45.4} \\ \Xhline{1.2pt}
\end{tabular}}
\caption{Macro P/R/F1 results on Ultra-Fine test set. * means using augmented data. "without label dependency" methods formulated FET as multi-label classification without considering associations between labels. "with label dependency" methods leveraged associations between labels explicitly or implicitly.}
\label{tab:Ultra-Fine main result}
\end{table}
\begin{table*}[tbh!]
\centering
\resizebox{.9\textwidth}{!}{
\begin{tabular}{l|cccccccccccc}
\Xhline{1.2pt}
\multicolumn{1}{c|}{\multirow{2}{*}{\textbf{Model}}} & \multicolumn{3}{c}{\textbf{Total}} & \multicolumn{3}{c}{\textbf{General}} & \multicolumn{3}{c}{\textbf{Fine}} & \multicolumn{3}{c}{\textbf{Ultra-Fine}} \\ \cline{2-13} 
\multicolumn{1}{c|}{} & \textbf{P} & \textbf{R} & \multicolumn{1}{c|}{\textbf{F}} & \textbf{P} & \textbf{R} & \multicolumn{1}{c|}{\textbf{F}} & \textbf{P} & \textbf{R} & \multicolumn{1}{c|}{\textbf{F}} & \textbf{P} & \textbf{R} & \textbf{F} \\ \hline
*\citet{choi2018ultra_data1} & 48.1 & 23.2 & \multicolumn{1}{c|}{31.3} & 60.3 & 61.6 & \multicolumn{1}{c|}{61.0} & 40.4 & 38.4 & \multicolumn{1}{c|}{39.4} & 42.8 & 8.8 & 14.6 \\
$\dagger$L\small{ABEL}\normalsize{GCN} \citep{xiong2019imposing_core2} & 49.3 & 28.1 & \multicolumn{1}{c|}{35.8} & 66.2 & 68.8 & \multicolumn{1}{c|}{67.5} & 43.9 & 40.7 & \multicolumn{1}{c|}{42.2} & 42.4 & 14.2 & 21.3 \\
HY Large \citep{DBLP:conf/emnlp/Lopez020_hyper} & 43.4 & 34.2 & \multicolumn{1}{c|}{38.2} & 61.4 & 73.9 & \multicolumn{1}{c|}{67.1} & 35.7 & 46.6 & \multicolumn{1}{c|}{40.4} & 36.5 & 19.9 & 25.7 \\
*ELMo \cite{DBLP:conf/naacl/OnoeD19_elmo} & 50.7 & 33.1 & \multicolumn{1}{c|}{40.1} & 66.9 & \textbf{80.7} & \multicolumn{1}{c|}{73.2} & 41.7 & 46.2 & \multicolumn{1}{c|}{43.8} & 45.6 & 17.4 & 25.2 \\
BERT \cite{DBLP:conf/naacl/OnoeD19_elmo} & 51.6 & 32.8 & \multicolumn{1}{c|}{40.1} & 67.4 & 80.6 & \multicolumn{1}{c|}{73.4} & 41.6 & 54.7 & \multicolumn{1}{c|}{47.3} & 46.3 & 15.6 & 23.4 \\ \hline
BERT[in-house] & 54.1 & 32.1 & \multicolumn{1}{c|}{40.3} & 68.8 & 79.2 & \multicolumn{1}{c|}{73.6} & 43.8 & \textbf{57.4} & \multicolumn{1}{c|}{49.7} & \textbf{50.7} & 14.6 & 22.6 \\
LRN \small w/o IR \normalsize & \textbf{60.7} & 32.5 & \multicolumn{1}{c|}{42.3} & \textbf{79.3} & 75.5 & \multicolumn{1}{c|}{\textbf{77.4}} & \textbf{59.6} & 44.8 & \multicolumn{1}{c|}{51.2} & 45.7 & 18.7 & 26.5 \\
LRN & 53.7 & \textbf{38.6} & \multicolumn{1}{c|}{\textbf{44.9}} & 77.8 & 76.4 & \multicolumn{1}{c|}{77.1} & 55.8 & 50.6 & \multicolumn{1}{c|}{\textbf{53.0}} & 43.4 & \textbf{26.0} & \textbf{32.5} \\ \Xhline{1.2pt}
\end{tabular}}
\caption{Macro P/R/F1 of each label granularity on Ultra-Fine dev set, and long tail labels are mostly in the ultra-fine layer. * means using augmented data. $\dagger$ We adapt the results from \citet{DBLP:conf/emnlp/Lopez020_hyper}.}
\label{tab:Ultra-Fine layer score}
\end{table*}
\begin{table*}[ht!]
\centering
\resizebox{.9\textwidth}{!}{
\begin{tabular}{l|cccccccccccc}
\Xhline{1.2pt}
\multicolumn{1}{c|}{\multirow{2}{*}{\textbf{Model}}} & \multicolumn{3}{c}{\textbf{Total}} & \multicolumn{3}{c}{\textbf{General}} & \multicolumn{3}{c}{\textbf{Fine}} & \multicolumn{3}{c}{\textbf{Ultra-Fine}} \\ \cline{2-13} 
\multicolumn{1}{c|}{} & \textbf{P} & \textbf{R} & \multicolumn{1}{c|}{\textbf{F}} & \textbf{P} & \textbf{R} & \multicolumn{1}{c|}{\textbf{F}} & \textbf{P} & \textbf{R} & \multicolumn{1}{c|}{\textbf{F}} & \textbf{P} & \textbf{R} & \textbf{F} \\ \hline
HY XLarge~\citep{DBLP:conf/emnlp/Lopez020_hyper} & / & / & \multicolumn{1}{c|}{/} & / & / & \multicolumn{1}{c|}{69.1} & / & / & \multicolumn{1}{c|}{39.7} & / & / & 26.1 \\
BERT[in-house] & 55.9 & 33.0 & \multicolumn{1}{c|}{41.5} & 69.7 & \textbf{81.6} & \multicolumn{1}{c|}{75.2} & 43.7 & \textbf{56.0} & \multicolumn{1}{c|}{49.1} & \textbf{53.5} & 15.5 & 24.0 \\
LRN \small w/o IR \normalsize & \textbf{61.2} & 33.5 & \multicolumn{1}{c|}{43.3} & \textbf{78.3} & 76.7 & \multicolumn{1}{c|}{\textbf{77.5}} & \textbf{61.6} & 44.1 & \multicolumn{1}{c|}{51.4} & 47.8 & 19.9 & 28.1 \\
LRN & 54.5 & \textbf{38.9} & \multicolumn{1}{c|}{\textbf{45.4}} & 77.4 & 76.7 & \multicolumn{1}{c|}{77.1} & 58.4 & 50.4 & \multicolumn{1}{c|}{\textbf{54.1}} & 43.5 & \textbf{26.4} & \textbf{32.8} \\ \Xhline{1.2pt}
\end{tabular}}
\caption{Macro P/R/F1 of different label granularity on Ultra-Fine test set.}
\label{tab:layer score on UF Test}
\end{table*}
\begin{table*}[thb!]
\setlength{\belowcaptionskip}{-0 cm}
\centering
\resizebox{.9\textwidth}{!}{
\begin{tabular}{l|c|c|ccc|ccc|ccc}
\Xhline{1.2pt}
\multicolumn{1}{c|}{\multirow{2}{*}{\textbf{Number of}}} & \multirow{2}{*}{\textbf{Category}} & \multirow{2}{*}{\textbf{Prediction}} & \multicolumn{3}{c|}{\textbf{Shot$=$0}} & \multicolumn{3}{c|}{\textbf{Shot$=$1}} & \multicolumn{3}{c}{\textbf{Shot$=$2}} \\ \cline{4-12} 
\multicolumn{1}{c|}{} &  &  & Correct & Predicted & Prec. &Correct & Predicted & Prec. & Correct & Predicted & Prec. \\ \hline
BERT[in-house] & 293 & 5683 & 0 & 0 & / & 1 & 1 & 100.0\% & 9 & 66 & 13.6\% \\ \cline{1-1}
LRN \small w/o IR \normalsize & 330 & 5740 & 0 & 0 & / & 1 & 3 & 33.3\% & 15 & 28 & 53.6\% \\ \cline{1-1}
LRN & 997 & 7808 & 110 & 218 & 50.5\% & 67 & 252 & 26.6\% & 94 & 276 & 34.1\% \\ \Xhline{1.2pt}
\end{tabular}}
\caption{Performance of the zero-shot, shot$=$1 and shot$=$2 label prediction. "Category" means how many kinds of types are predicted. "Prediction" means how many labels are generated.}
\label{tab:Ultra-Fine each shot}
\end{table*}
Table~\ref{tab:Ultra-Fine main result} shows the main results of all baselines and our method in two settings: LRN is the full model and LRN \small w/o IR \normalsize is the model without inductive reasoning. For fair comparisons, we implement a baseline with same settings of LRN but replace the decoder with a multi-classifier same as \citet{choi2018ultra_data1} --- BERT[in-house]. We can see that:

1) \textit{By performing label reasoning, LRN can effectively resolve the fine-grained entity typing problem.} Compared with previous methods, our method achieves state-of-the-art performance with a F1 improvement from 40.2 to 45.4 on test set. This verified the necessity for exploiting label dependencies for FET and the effectiveness of our two label reasoning mechanisms. We believe this is because label reasoning can help FET by making the learning more data-efficient (i.e., labels can share knowledge) and the prediction of labels global coherent.

2) \textit{Both deductive reasoning and inductive reasoning are useful for fine-grained label prediction.} Compared with BERT[in-house], LRN \small w/o IR \normalsize can achieve 4.3\% F1 improvement by exploiting extrinsic dependencies via deductive reasoning. LRN can further improve F1 from 43.3 to 45.4 by exploiting intrinsic dependencies via inductive reasoning. We believe this is because deductive reasoning and inductive reasoning are two fundamental but different mechanisms, therefore, modeling them simultaneously will better leverage label dependencies to predict labels.

3) \textit{Seq2Set is an effective framework to model, learn and exploit label dependencies in an end-to-end manner.} Compared with L\small{ABEL}\normalsize{GCN}~\citep{xiong2019imposing_core2} which heuristically exploits label co-occurrence structure, LRN can achieve a significant performance improvement. We believe this is because neural networks have strong ability for representing and learning label dependencies. And the end-to-end manner makes LRN can easily generalize to new scenarios.

\subsection{Effect on Long Tail Labels}
As described above, another advantage of our method is it can resolve the long tail problem by decomposing long tail labels to common attributes and modeling label dependencies between head and tail labels. Because the finer the label granularity, the more likely it to be a long tail label, we report the performance of each label granularity on dev set and test set same as previous works in Table~\ref{tab:Ultra-Fine layer score} and Table~\ref{tab:layer score on UF Test}. Moreover, we report the performance of the labels with shot$\leq$2 in Table~\ref{tab:Ultra-Fine each shot}. Based on these results, we find that:

1) \textit{LRN can effectively resolve the long tail label problem.} Compared to BERT[in-house], LRN can significantly improve the F-score of ultra-fine granularity labels by 44\% (22.6 $\rightarrow$ 32.5) and recall more fine-grained labels (14.6 $\rightarrow$ 26.0).

2) \textit{Both deductive reasoning and inductive reasoning are helpful for long tail label prediction, but with different underlying mechanisms: deductive reasoning exploits the extrinsic dependencies between labels, but inductive reasoning exploits the intrinsic dependencies between labels.} LRN \small w/o IR \normalsize cannot predict zero-shot labels because it resolves long tail labels by relating head labels with long tail labels, therefore it cannot predict unseen labels. By contrast, LRN can predict zero-shot labels via inductive reasoning because it can decompose labels into attributes. Furthermore, we found LRN \small w/o IR \normalsize has higher precision for few-shot (shot$=$2) labels than BERT and LRN, we believe this is because inductive reasoning focuses on recalling more labels, which inevitably introduce some incorrect labels.
\subsection{Detailed Analysis}
\paragraph{Effect of Components}
\begin{table}[htb!]
\centering
\resizebox{.45\textwidth}{!}{
\begin{tabular}{l|ccc|ccc}
\Xhline{1.2pt}
\multicolumn{1}{c|}{\multirow{2}{*}{\textbf{Model}}} & \multicolumn{3}{c|}{\textbf{Dev}} & \multicolumn{3}{c}{\textbf{Test}} \\ \cline{2-7} 
\multicolumn{1}{c|}{} & \textbf{P} & \textbf{R} & \textbf{F} & \textbf{P} & \textbf{R} & \textbf{F} \\ \hline
\textbf{LRN} & 53.7 & 38.6 & 44.9 & 54.5 & 38.9 & 45.4 \\
-PreAtt & 53.1 & 39.3 & 45.2 & 52.6 & 39.5 & 45.1 \\
-PreAtt-ConAtt & 56.3 & 36.3 & 44.2 & 56.4 & 36.5 & 44.3 \\
-SetLoss & 46.8 & 40.7 & 43.5 & 47.8 & 40.7 & 44.0 \\ \hline
\textbf{LRN \small w/o IR \normalsize} & 60.7 & 32.5 & 42.3 & 61.2 & 33.5 & 43.3 \\
-PreAtt & 54.5 & 34.2 & 42.1 & 55.1 & 35.0 & 42.8 \\
-PreAtt-ConAtt & 55.2 & 32.9 & 41.3 & 56.2 & 34.3 & 42.6 \\
-SetLoss & 46.0 & 37.6 & 41.4 & 46.6 & 37.5 & 41.6 \\ \Xhline{1.2pt}
\end{tabular}}
\caption{Ablation results on Ultra-Fine dataset: PreAtt denotes premise attention, ConAtt denotes contextual attention, and -SetLoss denotes replacing set prediction loss with cross-entropy loss.}
\label{Module Ablation}
\end{table}
\begin{figure}[htb!]
\setlength{\belowcaptionskip}{-0cm}
\centering
	\subfigure[]{
	\resizebox{0.23\textwidth}{!}{
   	 	\includegraphics{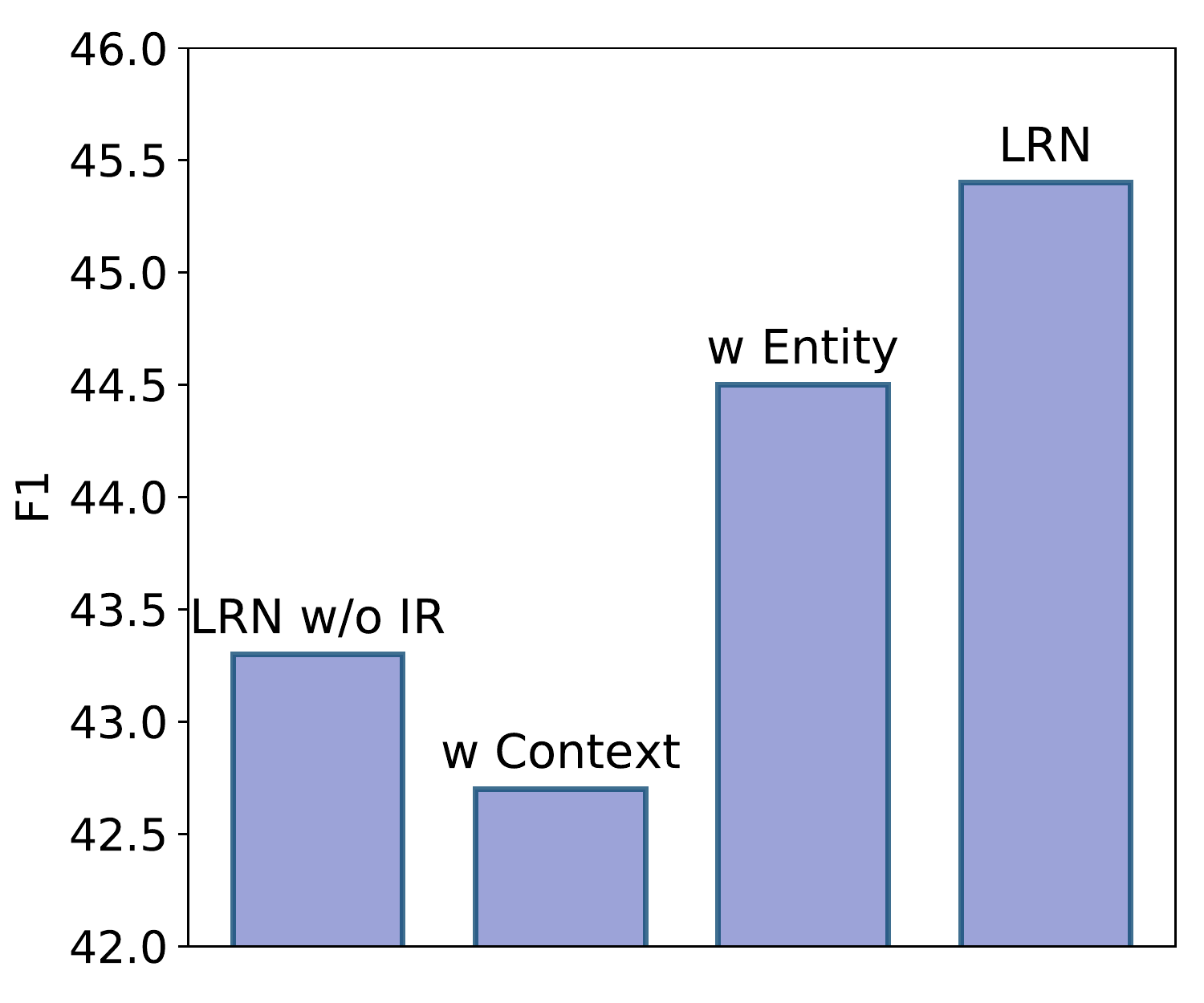}
	    \label{Fig.attribute}
	    }}
	\subfigure[]{
	\resizebox{0.23\textwidth}{!}{
			\includegraphics{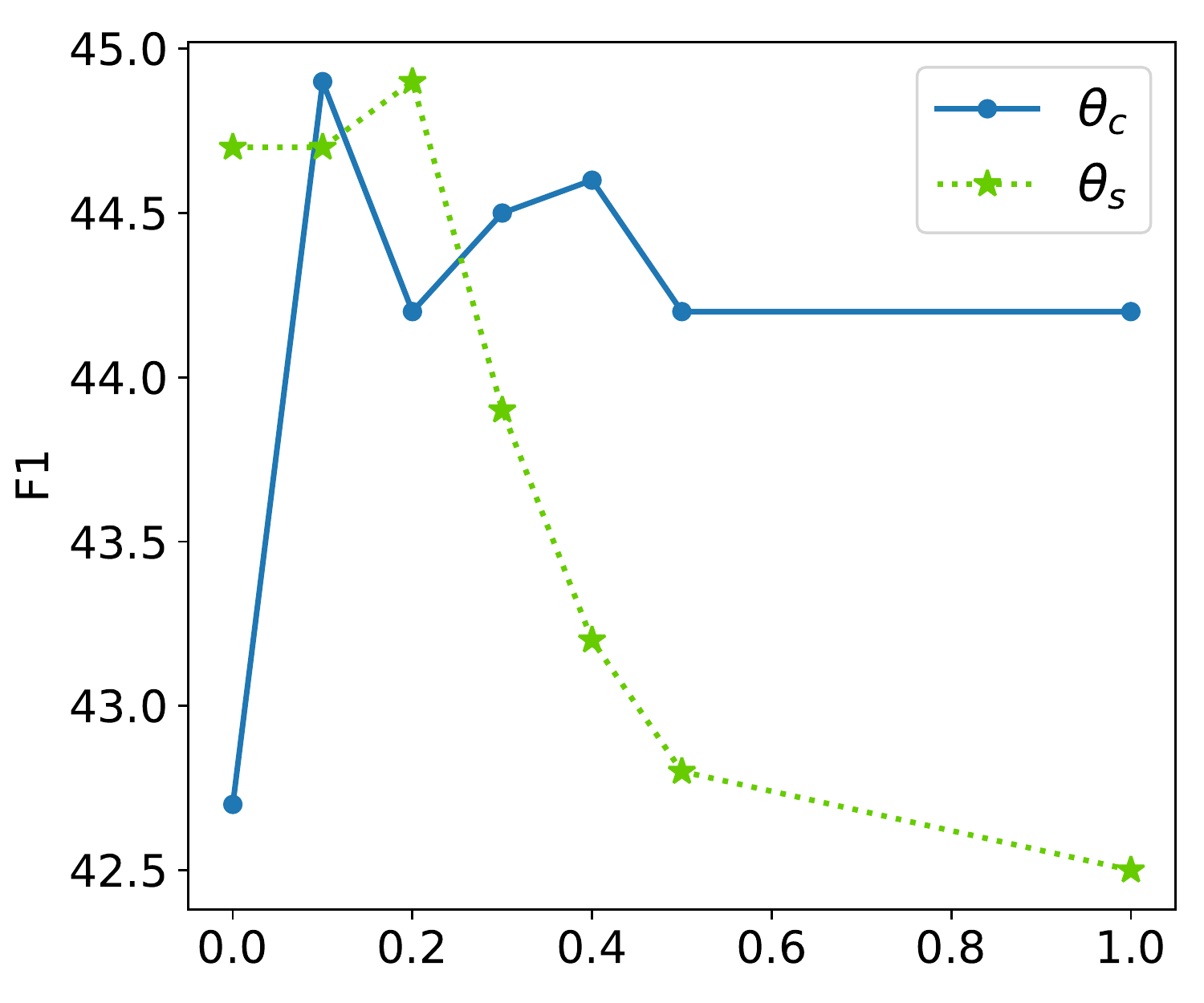}
		\label{Fig.thred}
	    }}
	\caption{(a) Ablation experiments of context attributes and entity attributes on Ultra-Fine dataset. (b) Performances of different confidence threshold ${\theta}_c$ and similarity threshold ${\theta}_s$ on dev set.} 
\end{figure}
To evaluate the effect of different components, we report the ablation results in Table~\ref{Module Ablation}. We can see that: (1) Set prediction loss is effective: replacing it with cross-entropy loss will lead to a significant decrease. (2) Both context and premise attention mechanisms are important for Seq2Set generation.
\paragraph{Effect of Attributes Set}
To explore the impact of entity attributes and context attributes in BAG, Figure~\ref{Fig.attribute} shows the results of different attributes configurations. We can see that: both attributes are useful, the context attribute has high coverage and may be noisy, while the entity attribute is opposite. However when introducing both of them, the information in entity attributes might help the context attributes to disambiguate them. This is similar to the effectiveness of contextual information in word sense disambiguation. As a result, these two kinds of attributes can complement each other. And Figure~\ref{Fig.thred} shows the performance on different thresholds, and we optimize confidence threshold ${\theta}_c=0.1$ and similarity threshold ${\theta}_s=0.2$ on dev set. Notice that ${\theta}_s$ is the threshold of activating labels and when ${\theta}_s=1$, it is equivalent to LRN \small w/o IR \normalsize.
\paragraph{Results of OntoNotes}
\begin{table}[!t]
\setlength{\belowcaptionskip}{-0cm}
\centering
\resizebox{.45\textwidth}{!}{
\begin{tabular}{llccc}
\Xhline{1.2pt}
\multicolumn{1}{c}{\textbf{Encoder}} & \multicolumn{1}{c|}{\textbf{Model}} & \textbf{Acc} & \textbf{MaF} & \textbf{MiF} \\ \Xhline{1.1pt}
\multicolumn{5}{c}{\textbf{with augmentation}} \\ \Xhline{1.1pt}
\multirow{1}{*}{HYPER} & \multicolumn{1}{l|}{\citet{DBLP:conf/emnlp/Lopez020_hyper}} & 47.4 & 75.8 & 69.4 \\ \hline 
\multirow{2}{*}{LSTM} & \multicolumn{1}{l|}{\citet{choi2018ultra_data1}} & 59.5 & 76.8 & 71.8 \\ 
 & \multicolumn{1}{l|}{\citet{xiong2019imposing_core2}} & 59.6 & 77.8 & 72.2 \\ \hline 
\multirow{2}{*}{ELMo} & \multicolumn{1}{l|}{*\citet{DBLP:conf/naacl/OnoeD19_elmo}} & 64.9 & 84.5 & 79.2 \\
 & \multicolumn{1}{l|}{\cite{linandJi2019attentive_core3}} & 63.8 & 82.9 & 77.3 \\ \hline
\multirow{4}{*}{BERT} & \multicolumn{1}{l|}{\citet{wang2020empirical}} & 61.1 & 81.8 & 76.3 \\
 & \multicolumn{1}{l|}{BERT {[}in-house{]}} & 62.2 & 83.4 & 78.8 \\
 & \multicolumn{1}{l|}{LRN \small w/o IR \normalsize} & \textbf{66.1} & \textbf{84.8} & \textbf{80.1} \\
 & \multicolumn{1}{l|}{LRN} & 64.5 & 84.5 & 79.3 \\ \Xhline{1.1pt}
\multicolumn{5}{c}{\textbf{without augmentation}} \\ \Xhline{1.1pt}
\multirow{2}{*}{ELMo} & \multicolumn{1}{l|}{*\citet{DBLP:conf/naacl/OnoeD19_elmo}} & 42.7 & 72.7 & 66.7 \\
 & \multicolumn{1}{l|}{\citet{DBLP:conf/acl/ChenCD20_onto_hier4}} & \textbf{58.7} & 73.0 & 68.1 \\ \hline
\multirow{4}{*}{BERT} & \multicolumn{1}{l|}{\citet{DBLP:conf/naacl/OnoeD19_elmo}} & 51.8 & 76.6 & 69.1 \\
 & \multicolumn{1}{l|}{BERT[in-house]} & 51.5 & 76.6 & 69.7 \\
 & \multicolumn{1}{l|}{LRN \small w/o IR \normalsize} & 55.3 & 77.3 & 70.4 \\
 & \multicolumn{1}{l|}{LRN} & 56.6 & \textbf{77.6} & \textbf{71.8} \\ \Xhline{1.2pt}
\end{tabular}}
\caption{Results on OntoNotes test set. Augmentation is the augmented data created by \cite{choi2018ultra_data1} which contains 800K instances and therefore there’re little few-shot labels in this setting. And * indicates using additional features to enhance the label representation.}
\label{ontonotes results}
\end{table}
To verify the generality of our method, we further conduct experiments on OntoNotes and report results of with and without augmentation data in Table~\ref{ontonotes results}. To embed labels in OntoNotes, we use the embedding of the last word of a label, e.g., \textit{/person/artist/director} is represented using embedding of \textit{director}.

We can see that: 1) LRN still achieves the best performance on both settings, which verified the robustness of our method. 2) Compared with Ultra-Fine, our method achieves a smaller improvement on OntoNotes. We found this is mainly because: First, OntoNotes has weaker label dependencies for its label set is smaller (89 vs 2519 for Ultra-Fine) and most of its labels are coarse-grained. Secondly, most labels in OntoNotes are frequent labels with many training instances, therefore the long tail label problem is not serious. This also explains why LRN \small w/o IR \normalsize can achieve better performance than LRN in the setting of with augmentation data: the more the training instance, the less need for long tail prediction.
\subsection{Case Study}
\begin{figure}[htb!]
\centering
\includegraphics[width=0.45\textwidth]{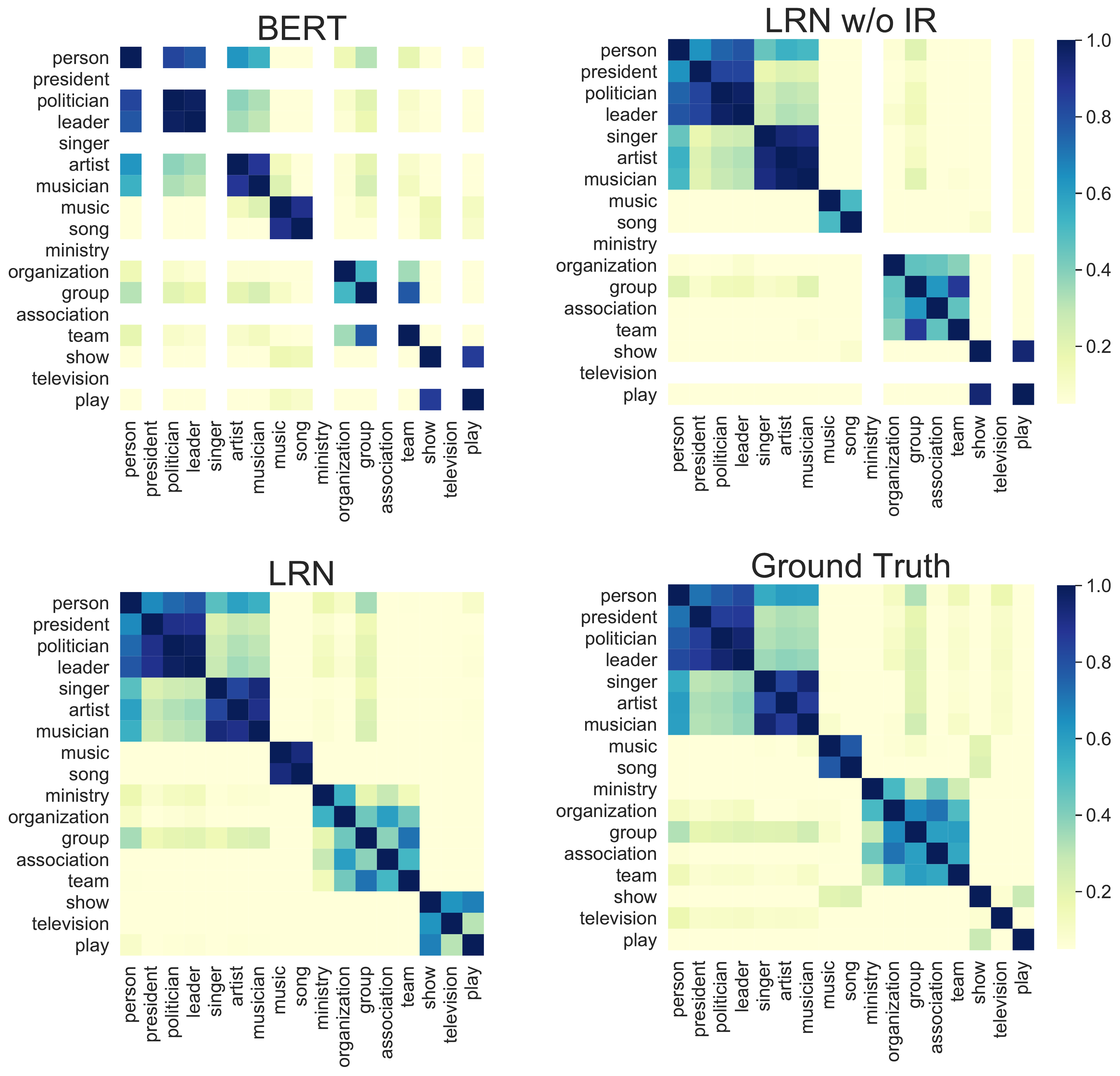}
\caption{Heat map of co-occurrence matrices of different models' prediction and ground truth. LRN w/o IR and LRN learn very similar co-occurrence matrices to Ground Truth.}
\label{Fig.heatmap}
\end{figure}
\begin{figure}[htb!]
\setlength{\belowcaptionskip}{-0cm}
\centering
\includegraphics[width=0.48\textwidth]{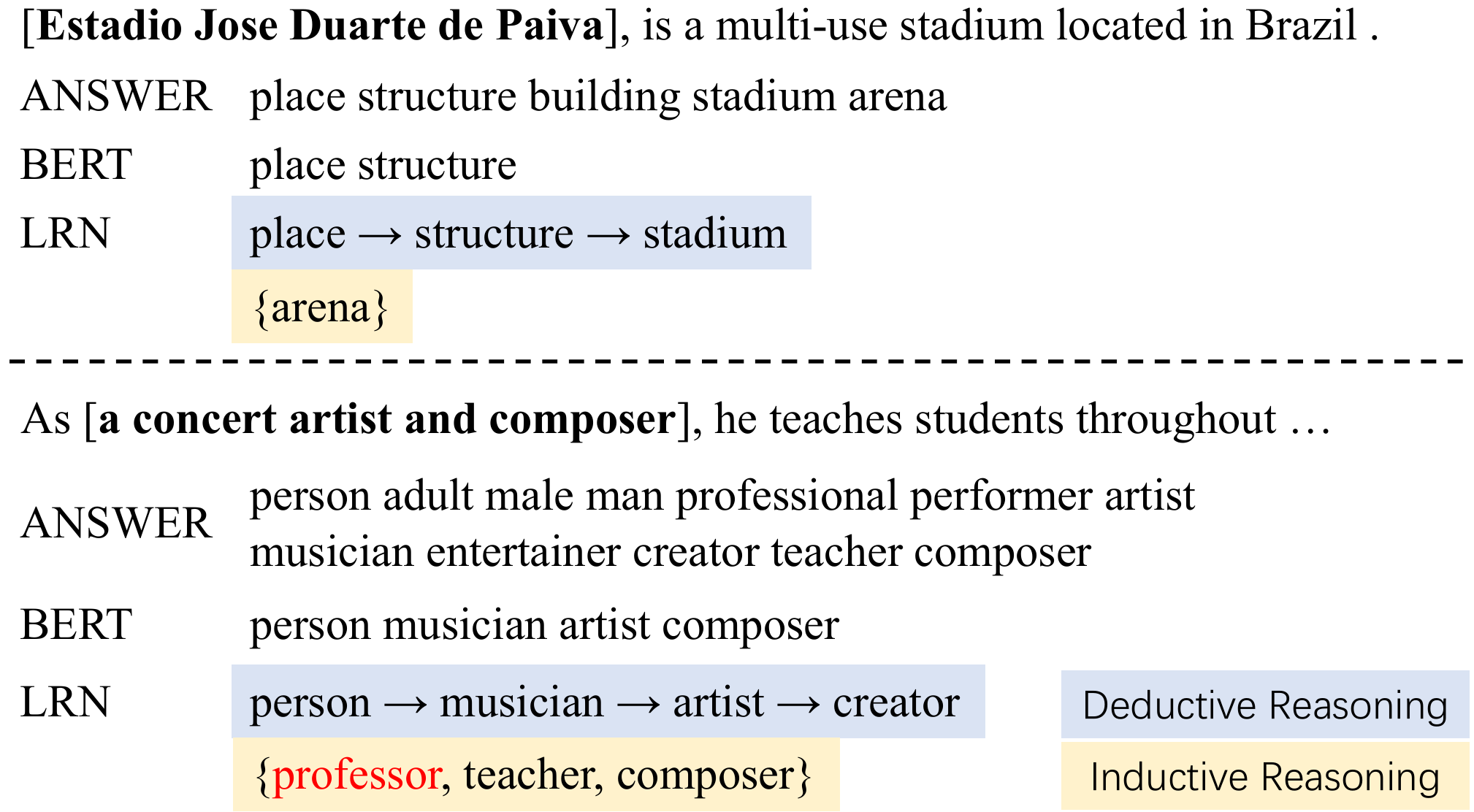} 
\caption{Cases of prediction results.}
\label{Fig.case} 
\end{figure}
To intuitively present the learned label dependencies, Figure~\ref{Fig.heatmap} shows the label co-occurrence matrices of different models' predictions and ground truth, we can see that both LRN and LRN \small w/o IR \normalsize can accurately learn label dependencies. Figure~\ref{Fig.case} shows some prediction cases and demonstrates that deductive and inductive reasoning have quite different underlying mechanisms and predict quite different labels.
\section{Conclusions}
This paper proposes \textit{Label Reasoning Network}, which uniformly models, learns and reasons complex label dependencies in a sequence-to-set, end-to-end manner. LRN designs two label reasoning mechanisms for effective decoding -- deductive reasoning to exploit extrinsic dependencies and inductive reasoning to exploit intrinsic dependencies. Experiments show that LRN can effectively cope with the massive label set on FET. And because our method uses no predefined structures, it can be easily generalized to new datasets and applied to other multi-classification tasks.
\section{Acknowledgments}
This work is supported by the National Key Research and Development Program of China (No. 2020AAA0106400), the National Natural Science Foundation of China under Grants no.U1936207 and 62106251, Beijing Academy of Artificial Intelligence (BAAI2019QN0502), and in part by the Youth Innovation Promotion Association CAS(2018141).
\bibliographystyle{acl_natbib}
\bibliography{emnlp2021.bib,anthology.bib}

\label{sec:appendix}
\label{sec:supplemental}

\end{document}